\documentclass{opt2025} 

\usepackage{multirow}
\usepackage{adjustbox}
\usepackage{framed}
\usepackage{wrapfig}
\usepackage{booktabs}
\newtheorem{assumption}[theorem]{Assumption}
\usepackage{framed}
\usepackage{caption}

\title[Gradient Autoscaled Normalization]{Insights from Gradient Dynamics: Gradient Autoscaled Normalization}

\optauthor{%
\Name{Vincent-Daniel Yun} \Email{juyoung.yun@usc.edu}\\
\addr University of Southern California, USA}

\begin{document}

\maketitle

\begin{abstract}%
Gradient dynamics play a central role in determining the stability and generalization of deep neural networks. In this work, we provide an empirical analysis of how variance and standard deviation of gradients evolve during training, showing consistent changes across layers and at the global scale in convolutional networks. Motivated by these observations, we propose a hyperparameter-free gradient normalization method that aligns gradient scaling with their natural evolution. This approach prevents unintended amplification, stabilizes optimization, and preserves convergence guarantees. Experiments on the challenging CIFAR-100 benchmark with ResNet-20, ResNet-56, and VGG-16-BN demonstrate that our method maintains or improves test accuracy even under strong generalization. Beyond practical performance, our study highlights the importance of directly tracking gradient dynamics, aiming to bridge the gap between theoretical expectations and empirical behaviors, and to provide insights for future optimization research.

\end{abstract}

\section{Introduction}
Gradient-based optimization is the foundation of modern deep learning. While stochastic gradient descent (SGD)~\cite{sgd} and its variants~\cite{sgd, kingma2015adam, adamw, center, gradnorm, yun2024znorm} have achieved remarkable success, the dynamics of gradients during training remain a central factor influencing both convergence speed and generalization~\cite{gen1, gen2, gen3, gen4}. Understanding these dynamics is particularly critical, as challenges such as vanishing and exploding gradients~\cite{gradvan1, gradvan2, gradvan3} are still active research topics and directly impact the stability and efficiency of training.

In this work, we investigate the gradient dynamics of convolutional neural networks (CNNs)~\cite{he2016deep, simonyan2015very}. Through empirical observation, we track the variance and standard deviation of gradients over training and find that layer-wise statistics change significantly as optimization progresses. Although prior theory suggests that gradients and their variance decrease during training~\cite{var1, ba1, ba2}, we directly track both layer-wise and global standard deviation, bridging the gap between theoretical expectation and empirical behavior. 

Motivated by these observations, we propose a hyperparameter-free gradient normalization method. Our approach adjusts gradient magnitudes to follow their natural evolution, ensuring they gradually diminish rather than being amplified, which stabilizes optimization and maintains consistent training dynamics. By explicitly visualizing how gradient statistics evolve, we aim to provide insights into their role in convergence and generalization, a direction that has received limited attention in prior work, especially in terms of tracking layer-wise and global gradients over time. This lays groundwork for more reliable optimization strategies and ultimately provides \textit{insights for future research}. The related works is provided in Appendix~\ref{related}.

\section{Problem Setup}
\subsection{Layer-wise Gradients Dynamics}
\begin{figure*}[h]
\centering
\includegraphics[width=1\columnwidth]{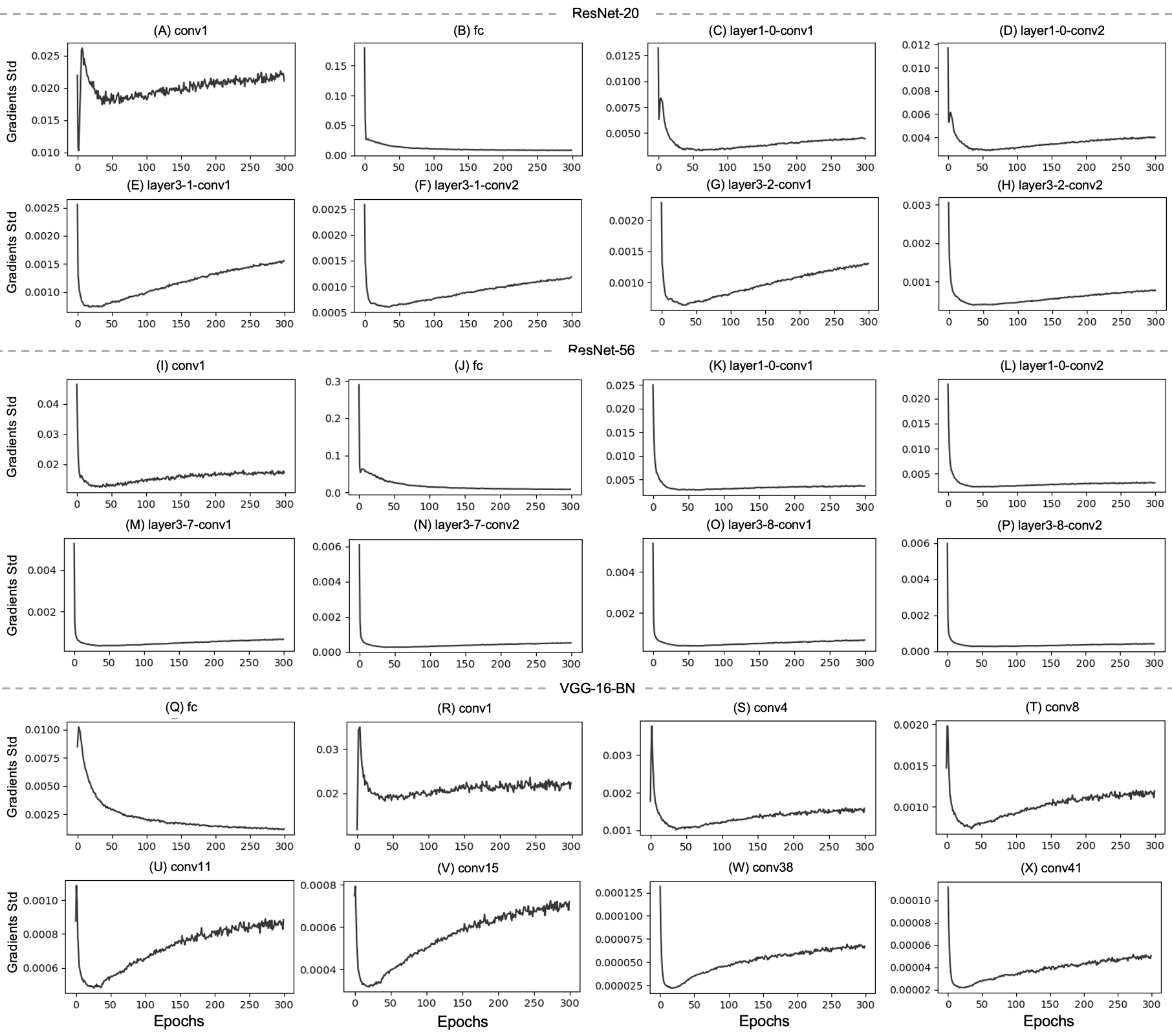}
\caption{Layer-wise Gradients Standard Deviation for ResNet-20~\cite{he2016deep}, ResNet-56~\cite{he2016deep}, and VGG-16-BN~\cite{simonyan2015very} Across 300 Epochs on CIFAR-10~\cite{krizhevsky2009learning}. All other settings are same with experimental results section~\ref{exp}}
\label{fig:overview}
\vspace{-10pt}
\end{figure*}

Understanding gradient dynamics is essential for optimizing deep neural networks, as shown in Figure~\ref{fig:overview}, which illustrates layer-wise gradient standard deviation across epochs for ResNet-20~\cite{he2016deep}, ResNet-56~\cite{he2016deep}, and VGG-16-BN~\cite{simonyan2015very} on CIFAR-10~\cite{krizhevsky2009learning}. We observed that gradient std varies across layers, decreasing in some layers while increasing in others, with the pattern depending on the model's architecture.

This variability poses a challenge: while some layers’ std decreases, others increase or fluctuate irregularly. As a result, when applying layer-wise normalization methods such as ZNorm~\cite{yun2024znorm}, certain layers experience unwanted amplification while others are suppressed, leading to inconsistent update magnitudes across the network. In particular, applying ZNorm $\frac{\nabla\mathcal{L}(\theta) - \nabla\mathcal{L}(\theta)_{mean}}{\nabla\mathcal{L}(\theta)_{std}}$ amplifies gradients in layers with low std, especially in architectures such as VGG, where small gradient values cause severe scaling by $1/\text{std}$, resulting in performance degradation or even divergence.

\subsection{Global Gradients Dynamics}
\begin{figure*}[h]
\centering
\includegraphics[width=1\columnwidth]{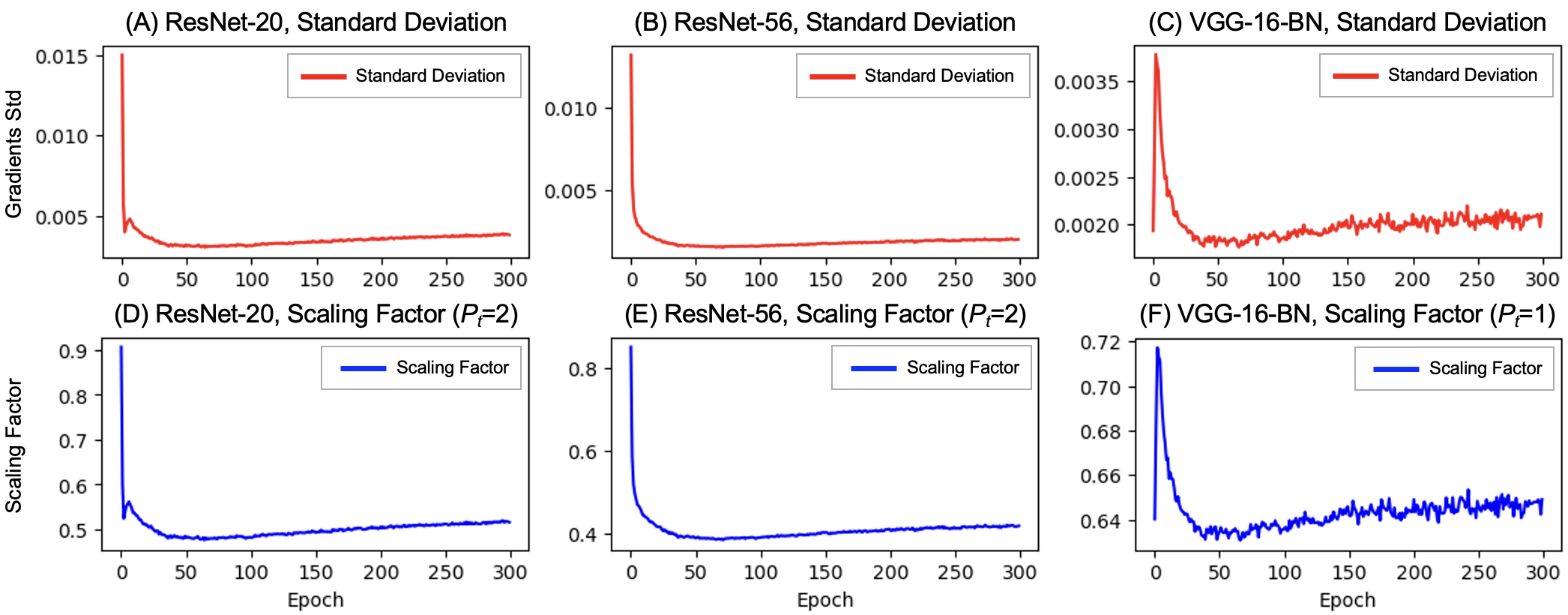}
\caption{Global gradients standard deviation (top) and corresponding scaling factor (bottom) for ResNet-20~\cite{he2016deep}, ResNet-56~\cite{he2016deep}, and VGG-16-BN~\cite{simonyan2015very} across 300 epochs on CIFAR-10~\cite{krizhevsky2009learning}. All other settings are same with experimental results section~\ref{exp} The exponent $P_t$ controls the curvature of the scaling map and is automatically determined in a hyperparameter-free setting, as explained later.}
\label{fig:global}
\vspace{-10pt}
\end{figure*}

The global gradient standard deviation provides a comprehensive view of the gradient dynamics across the entire network, as illustrated in Figure~\ref{fig:global}. This figure presents the evolution of global std over epochs for ResNet-20~\cite{he2016deep}, ResNet-56~\cite{he2016deep}, and VGG-16-BN~\cite{simonyan2015very} architectures. 

Unlike layer-wise analysis, where the std of individual layers may exhibit increases or irregular variations depending on the model's architecture, the global std demonstrates a consistent decreasing trend as training progresses. As epochs advance, it gradually declines, indicating a stabilization of the overall gradient distribution across all layers. This global perspective highlights a unifying trend that contrasts with the heterogeneous layer-wise behaviors, underscoring the importance of considering \texttt{network-wide dynamics} in optimization strategies.

Leveraging the observation that global gradients standard deviation decreases as training progresses, we have devised a method to normalize gradients using a scaling factor that diminishes over time, which will be detailed in the following methodology section.

\section{Methodology}
\subsection{Preliminaries}
We consider a deep neural network with $L$ layers and parameters $\{\boldsymbol{\theta}^{(l)}\}_{l=1}^L$. For fully connected layers, $\boldsymbol{\theta}^{(l)} \in \mathbb{R}^{D_l \times M_l}$; for convolutional layers, $\boldsymbol{\theta}^{(l)} \in \mathbb{R}^{C^{(l)}_{\mathrm{out}} \times C^{(l)}_{\mathrm{in}} \times k^{(l)}_1 \times k^{(l)}_2}$. Let $G^{(l)}_t \!:=\! \nabla_{\boldsymbol{\theta}^{(l)}} \mathcal{L}(\boldsymbol{\theta}_t)$ denote the gradient tensor of layer $l$ at iteration $t$. We denote by $\operatorname{vec}(\cdot)$ the vectorization operator and by
$\mu^{(l)}_t \;:=\; \frac{1}{n^{(l)}} \sum_{i=1}^{n^{(l)}} \big(G^{(l)}_t\big)_i$ the (entrywise) mean of $G^{(l)}_t$, where $n^{(l)}$ is the number of elements in $G^{(l)}_t$. We write $\operatorname{Std}(\cdot)$ for the (unbiased) standard deviation of a vector and use $\epsilon>0$ as a fixed numerical stabilizer (we use $\epsilon=10^{-8}$ in all experiments).

\subsection{Gradient Autocaled Normalization}
\label{sec:our proposed method}
Gradient Gradient Autoscaled Normalization performs a two-stage, statistics-driven modification of gradients before the standard gradient descent update~\cite{sgd}:
(i) \emph{layer-wise mean removal} (zero-centering)~\cite{center}, and
(ii) a \emph{global autoscale} multiplier shared by all eligible tensors at iteration $t$.
Unlike Z-score normalization~\cite{yun2024znorm}, Our proposed method does not divide by the (local) standard deviation, thereby avoiding uncontrolled amplification in layers whose gradient variance becomes very small.

\paragraph{Eligible set.}
At each iteration $t$, we form the set of tensors whose gradients have at least two dimensions,
$\mathcal{S}_t \;:=\; \big\{\, l \in \{1,\dots,L\} \;:\; \mathrm{dim}(G^{(l)}_t) > 1 \,\big\},$ which typically excludes bias vectors and scalar parameters. We aggregate all eligible gradients into a single vector
$\mathbf{g}_t \;:=\; \bigoplus_{l \in \mathcal{S}_t} \operatorname{vec}\!\big(G^{(l)}_t\big),$ and define the \emph{global} gradient standard deviation $s_t = \operatorname{Std}(\mathbf{g}_t)$

\begin{figure*}[h]
\centering
\includegraphics[width=1\columnwidth]{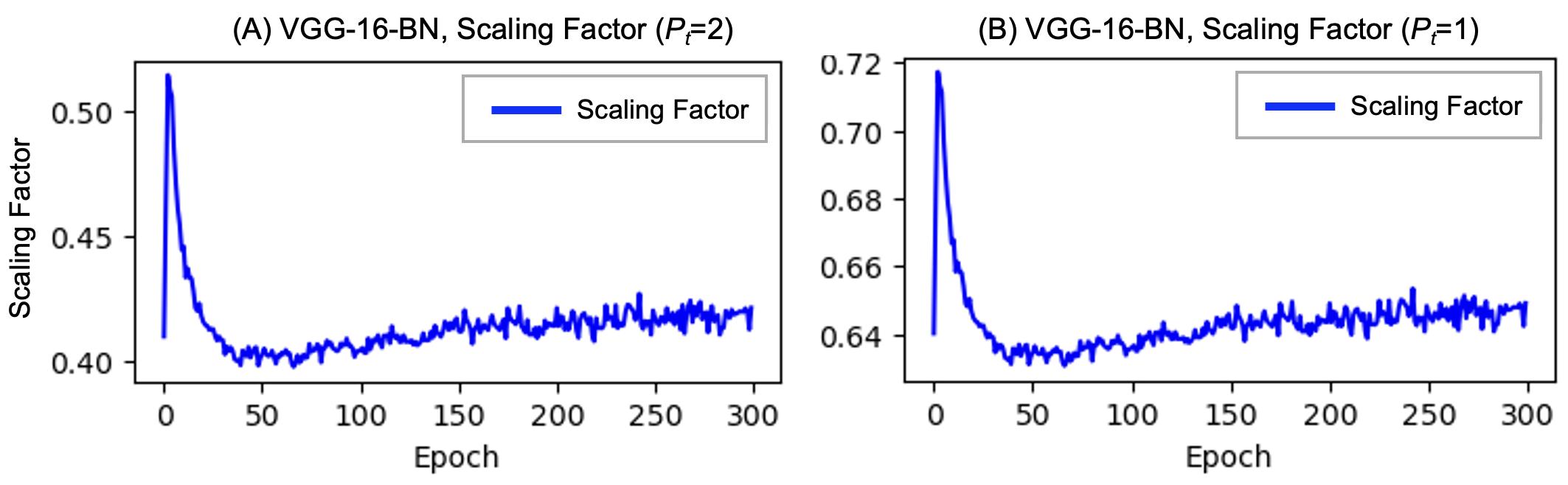}
\caption{Scaling factor on VGG-16-BN~\cite{simonyan2015very} with different settings: (A) $P_t=2$, (B) $P_t=1$.}
\label{fig:scaling_factor}
\vspace{-10pt}
\end{figure*}

\paragraph{Autoscale via log–std.}
Our proposed method maps the global scale $s_t$ to a scalar multiplier $a_t \in (0,+\infty)$ through a smooth, hyperparameter-free transform of $\lvert \log s_t\rvert$:
\begin{align}
a_t \;=\; \left(\left(0.5 + \frac{1}{0.5\left(\lvert \log s_t\rvert + \epsilon\right)}\right) - 0.5\right)\cdot 2^{\,p_t},
\quad
p_t \;=\;
\begin{cases}
1, & \text{for all } t \ge 2 \text{ if } a_1 < 0.5, \\
2, & \text{otherwise}. \label{eq1}
\end{cases}
\end{align}
Algebraically, eq~\ref{eq1} simplifies to $a_t = \left(\frac{4}{\lvert \log s_t\rvert + \epsilon}\right)^{p_t}$ where $p_t$ is an integer \emph{exponent} that controls the curvature of the scaling map. Figure~\ref{fig:scaling_factor} illustrates the behavior of the scaling factor for VGG-16-BN when $p_t=2$ (left) and $p_t=1$ (right). When the initial standard deviation $s_1$ is small, the quadratic mapping ($p_t=2$) can reduce $a_t$ excessively, leading to overly aggressive downscaling. To prevent this, our proposed method adopts a one-shot safeguard: the exponent $p_t$ is chosen adaptively at the first iteration according to eq~\ref{eq1}.

This rule ensures that the scaling factor remains stable and avoids excessive shrinkage in the presence of anomalously small initial gradient variance, while still allowing curvature control in subsequent epochs.

\paragraph{Layer-wise zero-centering and global rescaling.}
For each eligible layer $l \in \mathcal{S}_t$, we compute the mean-removed gradient $\widetilde{G}^{(l)}_t = G^{(l)}_t - \mu^{(l)}_t \cdot \mathbf{1}$ where $\mathbf{1}$ is the all-ones tensor of the same shape as $G^{(l)}_t$. our proposed method then applies the \emph{same} autoscale $a_t$ to all eligible layers:
\begin{equation}
\label{eq:global_rescale}
\widehat{G}^{(l)}_t \;=\; a_t \cdot \widetilde{G}^{(l)}_t
\quad \text{for all } l \in \mathcal{S}_t,
\qquad
\widehat{G}^{(l)}_t \;=\; G^{(l)}_t
\quad \text{for } l \notin \mathcal{S}_t.
\end{equation}
The above equations describe a \emph{zero-mean, globally consistent} gradient transformation that avoids dividing by very small per-layer standard deviations, preventing gradient explosion at the layer level while still adapting to the overall gradient scale of the network.

\paragraph{SGD update.}
Let $\widehat{\mathbf{g}}_t := \bigoplus_{l=1}^L \operatorname{vec}\!\big(\widehat{G}^{(l)}_t\big)$ denote the transformed gradient concatenated over all parameters. our proposed method with standard stochastic gradient descent (SGD)~\cite{sgd} performs the following update at iteration $t$: $\boldsymbol{\theta}_{t+1}=\boldsymbol{\theta}_t-\eta \cdot \widehat{\mathbf{g}}_t$ where $\eta > 0$ is the learning rate. This formulation highlights that the effect of our proposed method lies entirely in the gradient transformation, while the underlying optimizer step remains identical to standard SGD~\cite{sgd}. This allows easy adaptation to other optimizers such as Adam~\cite{kingma2015adam, adamw} and RMSProp~\cite{tieleman2012rmsprop}.


\section{Theoretical Analysis: Convergence
Guarantees}
\begin{lemma}
Suppose the loss function \(L\) is \(\beta\)-smooth and the stochastic gradient is unbiased with bounded variance, i.e., \(\mathbb{E}[\nabla L_t(w_t)] = \nabla L(w_t)\) and \(\mathbb{E}[\|\nabla L_t(w_t) - \nabla L(w_t)\|^2] \leq \tfrac{\sigma^2}{b}\). If the effective step size satisfies \(\eta a_t \leq \tfrac{1}{\beta}\), then the scaled SGD update \(w_{t+1} = w_t - \eta a_t \nabla L_t(w_t)\) guarantees:
\begin{align}
\mathbb{E}[L(w_{t+1})] \leq \mathbb{E}[L(w_t)] - \tfrac{\eta a_t}{2} \mathbb{E}[\|\nabla L(w_t)\|^2] + \tfrac{(\eta a_t)^2 \beta \sigma^2}{2b}. 
\end{align}
\end{lemma}

\begin{theorem}
Let the loss function \(L\) be \(\beta\)-smooth, and assume the stochastic gradient is unbiased with bounded variance, i.e., \(\mathbb{E}[\nabla L_t(w_t)] = \nabla L(w_t)\) and \(\mathbb{E}[\|\nabla L_t(w_t) - \nabla L(w_t)\|^2] \leq \tfrac{\sigma^2}{b}\). If the effective step size satisfies \(\eta a_t \leq \tfrac{1}{\beta}\), then the scaled SGD update \(w_{t+1} = w_t - \eta a_t \nabla L_t(w_t)\) guarantees:
\begin{align}
\frac{1}{T} \sum_{t=0}^{T-1} \mathbb{E}[\|\nabla L(w_t)\|^2] \leq \frac{2}{T \eta a_t} \big( L(w_0) - \mathbb{E}[L(w_T)] \big) + \frac{\eta a_t \beta \sigma^2}{b}.
\end{align}
\end{theorem}

\noindent
The complete proofs follow the standard analysis of SGD~\cite{sgd} and are deferred to Appendix~\ref{proof}. Our contribution is the inclusion of the scaling factor \(a_t\) in the update rule. Since \(a_t \in (0,1]\), it simply reduces the effective stepsize while leaving the overall convergence guarantees maintained.

\section{Experimental Results}
\label{exp}
\noindent \textbf{Experimental Settings.} Experiments are conducted on CIFAR-100~\cite{krizhevsky2009learning}, which poses a more challenging benchmark than CIFAR-10. We utilized GitHub repository~\cite{vit-cifar}, all models are trained without pretrained weights. We use AdamW~\cite{kingma2015adam,adamw} (lr=0.001, weight decay $5\times10^{-5}$) with step decay (0.75 every 30 epochs), batch size 256, and 200 epochs. Backbones include VGG-16-BN~\cite{simonyan2015very}, ResNet-20~\cite{he2016deep}, and ResNet-56~\cite{he2016deep}. For fair comparison with strong base, label smoothing~\cite{label} and CutMix~\cite{cutmix} are used. These generalization techniques create a setting where achieving performance gains is inherently difficult, underscoring the robustness of the proposed method. \\

\begin{table*}[h]
\centering
\begin{minipage}{0.48\textwidth}
\tiny
\begin{adjustbox}{width=\textwidth}
\begin{tabular}{llllllll}
\toprule
Network & Method & Top-1 Test Acc. \\
\midrule
\multirow{5}{*}{ResNet-20~\cite{he2016deep}} & AdamW (Baseline)~\cite{adamw} & 0.5932 \\
& AdamW + Gradient Normalization~\cite{gradnorm} & 0.5958 \\
& AdamW + Gradient Centralization~\cite{center} & 0.6072 \\
& AdamW + Z-Score Normalization~\cite{yun2024znorm} & 0.5953 \\
& AdamW + Ours & \textbf{0.6134} \\
\midrule
\multirow{5}{*}{ResNet-56~\cite{he2016deep}} & AdamW (Baseline)~\cite{adamw} & 0.7001 \\
& AdamW + Gradient Normalization~\cite{gradnorm} & 0.7050 \\
& AdamW + Gradient Centralization~\cite{center} & 0.6973 \\
& AdamW + Z-Score Normalization~\cite{yun2024znorm} & 0.6858 \\
& AdamW + Ours & \textbf{0.7129} \\
\midrule
\multirow{5}{*}{VGG-16-BN~\cite{simonyan2015very}} & AdamW (Baseline)~\cite{adamw} & \textbf{0.7454} \\
& AdamW + Gradient Normalization~\cite{gradnorm} & 0.7418 \\
& AdamW + Gradient Centralization~\cite{center} & 0.7382 \\
& AdamW + Z-Score Normalization~\cite{yun2024znorm} & 0.7391 \\
& AdamW + Ours & \textbf{0.7454} \\
\bottomrule
\end{tabular}
\end{adjustbox}
\caption{Comparison of performance on CIFAR-100~\cite{krizhevsky2009learning} across ResNet-20~\cite{he2016deep}, ResNet-56~\cite{he2016deep}, and VGG-16-BN~\cite{simonyan2015very}. Reported results include Top-1 test accuracy}
\label{table:exp_all}
\end{minipage}%
\hfill
\begin{minipage}{0.48\textwidth}
\centering
\includegraphics[width=\linewidth]{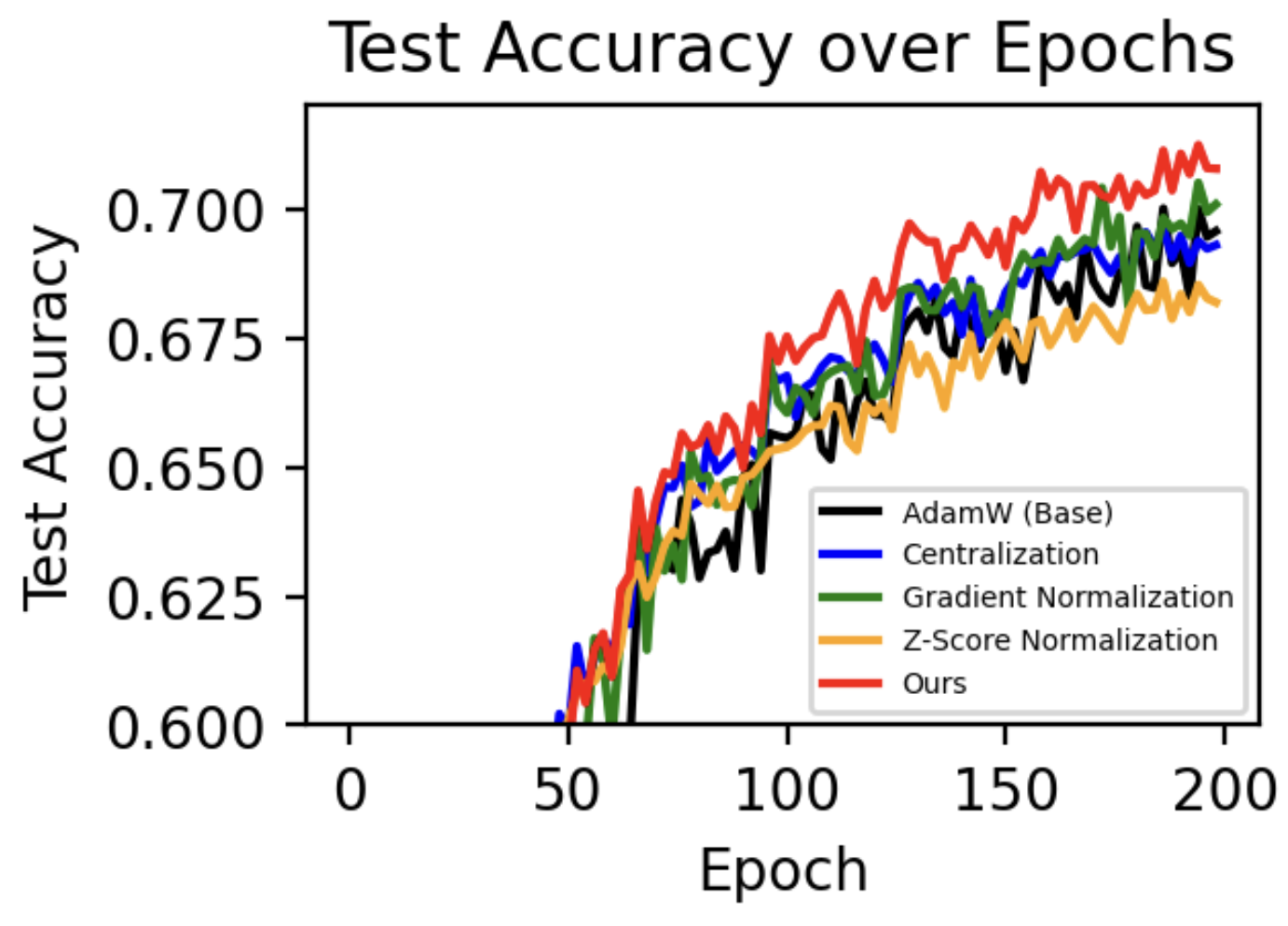}
\captionof{figure}{Top-1 Test accuracy comparison on ResNet-56~\cite{he2016deep}}
\label{fig:intro2}
\end{minipage}
\end{table*}

Our method consistently improves accuracy on ResNet architectures, while on VGG it achieves performance comparable to the baseline. Notably, alternative normalization techniques often lead to degraded results, underscoring the stability of the proposed approach.

\section{Discussion}
Our work analyzed gradient dynamics in convolutional networks such as ResNet and VGG, demonstrating the effect of hyperparameter-free normalization. As shown in Figure~\ref{fig:intro2}, our method consistently achieves higher test accuracy on ResNet-56~\cite{he2016deep} and exhibits smoother convergence compared to baseline AdamW~\cite{adamw} and other normalization approaches. This highlights the benefit of aligning gradient scaling with the natural evolution of gradient statistics. While these results validate the method, they also reveal limitations: the observations are tied to CNNs~\cite{he2016deep, simonyan2015very} and experiments were restricted to CIFAR-100~\cite{krizhevsky2009learning}. Future research will extend this analysis to Vision Transformers, whose gradient dynamics differ fundamentally due to attention-based architectures.

\section{Conclusion}
In this work, we analyzed the gradient dynamics of CNN and proposed a hyperparameter-free normalization method. By aligning gradient scaling with the natural evolution of gradient statistics, our approach mitigates uncontrolled amplification and improves generalization. Experimental results on CIFAR-100 confirm the robustness of our method even under strong generalization settings. We hope this study not only contributes a practical optimization technique but also provides insights into the role of gradient dynamics, paving the way for more robust training strategies.

\bibliography{sample}
\newpage
\appendix
\renewcommand{\thetheorem}{\arabic{theorem}}
\renewcommand{\thelemma}{\arabic{lemma}}
\setcounter{theorem}{0}

\section{Related Works}
\label{related}
Research on gradient-based optimization has mainly focused on how training dynamics influence both convergence and generalization. Early studies showed the importance of stochastic gradient descent~\cite{sgd} and its variants~\cite{adamw,kingma2015adam, tieleman2012rmsprop} in large-scale learning problems~\cite{large, large2}, providing both theory and strong results in practice. Later work studied the problems of vanishing and exploding gradients~\cite{gradvan1, gradvan2, gradvan3, gradvan4}, which are still major challenges for stable training, especially in very deep or recurrent networks. These difficulties motivated stabilization techniques such as batch normalization~\cite{ba1, ba2}, layer normalization~\cite{la1}, and residual connections~\cite{he2016deep,res1, res2, res3}.

A complementary line of research proposed methods that normalize gradients directly. Gradient normalization methods such as GradNorm~\cite{gradnorm} were first introduced for multitask learning, to keep gradients from different tasks balanced. A common rule is $g_i \leftarrow \frac{g_i}{\|g_i\|} \cdot \alpha,$ where $g_i$ is the gradient of task $i$ and $\alpha$ is a scaling factor. Gradient clipping~\cite{gradvan1} is another widely used method, defined as $g \leftarrow \frac{g}{\max(1, \|g\|/c)},$ where $c$ is a threshold, which prevents exploding gradients by shrinking overly large updates. Gradient centralization~\cite{center} improved generalization by simply subtracting the mean, $g \leftarrow {g - \mu_g}$. More recent work, such as Z-Score Gradient Normalization~\cite{yun2024znorm}, tried to standardize gradients across layers, $g \leftarrow \frac{g - \mu_g}{\sigma_g + \epsilon},$ but this can become unstable when some layers have very small variance, leading to amplification and worse performance.

While prior methods mainly focus on manipulating gradients, few works have explicitly tracked how their standard deviation evolves across layers or at the global scale. This study directly monitors these dynamics during training, linking theoretical expectations of gradient decay with empirical behavior. Building on this, a hyperparameter-free approach using global statistics is introduced, which avoids unstable amplification and preserves convergence guarantees, while highlighting the importance of studying gradient dynamics themselves as a basis for future optimization research.

\newpage
\section{Theoretical Analysis: Convergence Guarantees on SGD}
\label{proof}

\begin{assumption}[$\beta$-smoothness]
\label{assumption:smoothness}
The loss function \(L:\mathbb{R}^d \to \mathbb{R}\) is \(\beta\)-smooth, i.e., its gradient is \(\beta\)-Lipschitz continuous $\|\nabla L(u) - \nabla L(v)\| \;\leq\; \beta \|u-v\|, \forall u,v \in \mathbb{R}^d.$
Equivalently, for any \(u,v \in \mathbb{R}^d\), the following descent lemma holds:
\begin{align}
L(u) \;\leq\; L(v) + \langle \nabla L(v), u-v \rangle + \tfrac{\beta}{2}\|u-v\|^2.
\end{align}
\end{assumption}

\begin{assumption}[Unbiased stochastic gradient]
\label{assumption:unbiased}
At each iteration \(t\), the stochastic gradient \(\nabla L_t(w_t)\) is an unbiased estimator of the true gradient:
\begin{align}
\mathbb{E}[\nabla L_t(w_t)] \;=\; \nabla L(w_t).
\end{align}
\end{assumption}

\noindent \textbf{Lemma 1 }
\textit{Suppose the loss function \(L\) is \(\beta\)-smooth and the stochastic gradient is unbiased with bounded variance, i.e., \(\mathbb{E}[\nabla L_t(w_t)] = \nabla L(w_t)\) and \(\mathbb{E}[\|\nabla L_t(w_t) - \nabla L(w_t)\|^2] \leq \tfrac{\sigma^2}{b}\). If the effective step size satisfies \(\eta a_t \leq \tfrac{1}{\beta}\), then the scaled SGD update \(w_{t+1} = w_t - \eta a_t \nabla L_t(w_t)\) guarantees:}
\begin{align}
\mathbb{E}[L(w_{t+1})] \leq \mathbb{E}[L(w_t)] - \tfrac{\eta a_t}{2} \mathbb{E}[\|\nabla L(w_t)\|^2] + \tfrac{(\eta a_t)^2 \beta \sigma^2}{2b}. 
\end{align}

\begin{proof}
By the \(\beta\)-smoothness of \(L\), it holds that
\begin{align}
L(w_{t+1}) \leq L(w_t) + \langle \nabla L(w_t), w_{t+1} - w_t \rangle + \tfrac{\beta}{2} \|w_{t+1} - w_t\|^2.
\end{align}
Plugging in the update rule \(w_{t+1} = w_t - \eta a_t \nabla L_t(w_t)\), we obtain
\begin{align}
L(w_{t+1}) \leq L(w_t) - \eta a_t \langle \nabla L(w_t), \nabla L_t(w_t) \rangle + \tfrac{(\eta a_t)^2 \beta}{2} \|\nabla L_t(w_t)\|^2. 
\end{align}
Taking expectations and using the condition \(\mathbb{E}[\nabla L_t(w_t)] = \nabla L(w_t)\), it follows that
\begin{align}
\mathbb{E}[L(w_{t+1})] &\leq \mathbb{E}[L(w_t)] - \eta a_t \mathbb{E}[\langle \nabla L(w_t), \nabla L_t(w_t) \rangle] + \tfrac{(\eta a_t)^2 \beta}{2} \mathbb{E}[\|\nabla L_t(w_t)\|^2]  \\
&= \mathbb{E}[L(w_t)] - \eta a_t \mathbb{E}[\|\nabla L(w_t)\|^2] + \tfrac{(\eta a_t)^2 \beta}{2} \mathbb{E}[\|\nabla L_t(w_t)\|^2].
\end{align}
For the variance term, observe that
\begin{align}
\mathbb{E}[\|\nabla L_t(w_t)\|^2] &= \mathbb{E}[\|\nabla L_t(w_t) - \nabla L(w_t) + \nabla L(w_t)\|^2]  \\
&= \mathbb{E}[\|\nabla L_t(w_t) - \nabla L(w_t)\|^2 + \|\nabla L(w_t)\|^2 \\
& \quad + 2 \langle \nabla L_t(w_t) - \nabla L(w_t), \nabla L(w_t) \rangle ]  \\
&= \mathbb{E}[\|\nabla L_t(w_t) - \nabla L(w_t)\|^2] + \mathbb{E}[\|\nabla L(w_t)\|^2] \\
&= \mathbb{E}[\|\nabla L_t(w_t) - \nabla L(w_t)\|^2 + \|\nabla L(w_t)\|^2 \\
& \quad + 2 \mathbb{E}[\langle \nabla L_t(w_t) - \nabla L(w_t), \nabla L(w_t) \rangle]  \\
&= \mathbb{E}[\|\nabla L_t(w_t) - \nabla L(w_t)\|^2] + \mathbb{E}[\|\nabla L(w_t)\|^2] \\
&= \mathbb{E}[\|\nabla L_t(w_t) - \nabla L(w_t)\|^2 + \|\nabla L(w_t)\|^2 \\
& \quad + 2 \mathbb{E}[\langle \nabla L_t(w_t), \nabla L(w_t) \rangle - \langle \nabla L(w_t), \nabla L(w_t) \rangle]  \\
&= \mathbb{E}[\|\nabla L_t(w_t) - \nabla L(w_t)\|^2] + \mathbb{E}[\|\nabla L(w_t)\|^2]. \label{eq23}
\end{align}
Here the cross term vanishes since \(\mathbb{E}[\nabla L_t(w_t) - \nabla L(w_t)] = 0\).  
By the bounded variance condition \(\mathbb{E}[\|\nabla L_t(w_t) - \nabla L(w_t)\|^2] \leq \tfrac{\sigma^2}{b}\), we have
\begin{align}
\mathbb{E}[\|\nabla L_t(w_t)\|^2] \leq \mathbb{E}[\|\nabla L(w_t)\|^2] + \tfrac{\sigma^2}{b}. \label{eq24}
\end{align}
Substituting (\ref{eq23}) into (\ref{eq24}), we arrive at
\begin{align}
\mathbb{E}[L(w_{t+1})] &\leq \mathbb{E}[L(w_t)] - \eta a_t \mathbb{E}[\|\nabla L(w_t)\|^2] + \tfrac{(\eta a_t)^2 \beta}{2} \left( \mathbb{E}[\|\nabla L(w_t)\|^2] + \tfrac{\sigma^2}{b} \right) \notag \\
&= \mathbb{E}[L(w_t)] - \eta a_t \mathbb{E}[\|\nabla L(w_t)\|^2] + \tfrac{(\eta a_t)^2 \beta}{2} \mathbb{E}[\|\nabla L(w_t)\|^2] + \tfrac{(\eta a_t)^2 \beta}{2} \cdot \tfrac{\sigma^2}{b} \notag \\
&= \mathbb{E}[L(w_t)] - \eta a_t \mathbb{E}[\|\nabla L(w_t)\|^2] + \tfrac{(\eta a_t)^2 \beta}{2} \mathbb{E}[\|\nabla L(w_t)\|^2] + \tfrac{(\eta a_t)^2 \beta \sigma^2}{2b} \notag \\
&= \mathbb{E}[L(w_t)] + \left( -\eta a_t + \tfrac{(\eta a_t)^2 \beta}{2} \right) \mathbb{E}[\|\nabla L(w_t)\|^2] + \tfrac{(\eta a_t)^2 \beta \sigma^2}{2b} \notag \\
&= \mathbb{E}[L(w_t)] + \left( \tfrac{(\eta a_t)^2 \beta}{2} - \eta a_t \right) \mathbb{E}[\|\nabla L(w_t)\|^2] + \tfrac{(\eta a_t)^2 \beta \sigma^2}{2b} \notag \\
&= \mathbb{E}[L(w_t)] - \left( \eta a_t - \tfrac{(\eta a_t)^2 \beta}{2} \right) \mathbb{E}[\|\nabla L(w_t)\|^2] + \tfrac{(\eta a_t)^2 \beta \sigma^2}{2b}.
\end{align}
Finally, using \(\eta a_t \leq \tfrac{1}{\beta}\), we have
\begin{align}
\eta a_t - \tfrac{(\eta a_t)^2 \beta}{2} = \eta a_t \left( 1 - \tfrac{\eta a_t \beta}{2} \right) \geq \tfrac{\eta a_t}{2}, 
\end{align}
since \(\eta a_t \beta \leq 1\) implies \(1 - \tfrac{\eta a_t \beta}{2} \geq \tfrac{1}{2}\). Therefore,
\begin{align}
\mathbb{E}[L(w_{t+1})] \leq \mathbb{E}[L(w_t)] - \tfrac{\eta a_t}{2} \mathbb{E}[\|\nabla L(w_t)\|^2] + \tfrac{(\eta a_t)^2 \beta \sigma^2}{2b}. 
\end{align}
\end{proof}

\noindent \textbf{Theorem 2 }
\textit{Let the loss function \(L\) be \(\beta\)-smooth, and assume the stochastic gradient is unbiased with bounded variance, i.e., \(\mathbb{E}[\nabla L_t(w_t)] = \nabla L(w_t)\) and \(\mathbb{E}[\|\nabla L_t(w_t) - \nabla L(w_t)\|^2] \leq \tfrac{\sigma^2}{b}\). If the effective step size satisfies \(\eta a_t \leq \tfrac{1}{\beta}\), then the scaled SGD update \(w_{t+1} = w_t - \eta a_t \nabla L_t(w_t)\) guarantees:}
\begin{align}
\frac{1}{T} \sum_{t=0}^{T-1} \mathbb{E}[\|\nabla L(w_t)\|^2] \leq \frac{2}{T \eta a_t} \big( L(w_0) - \mathbb{E}[L(w_T)] \big) + \frac{\eta a_t \beta \sigma^2}{b}.
\end{align}

\begin{proof}
From the descent lemma applied to scaled SGD, we have
\begin{align}
\mathbb{E}[L(w_{t+1})] \leq \mathbb{E}[L(w_t)] - \tfrac{\eta a_t}{2} \mathbb{E}[\|\nabla L(w_t)\|^2] + \tfrac{(\eta a_t)^2 \beta \sigma^2}{2b}. \label{eq29}
\end{align}
Averaging inequality (\ref{eq29}) over \(T\) iterations yields
\begin{align}
\frac{1}{T} \sum_{t=0}^{T-1} \mathbb{E}[L(w_{t+1})] &\leq \frac{1}{T} \sum_{t=0}^{T-1} \left( \mathbb{E}[L(w_t)] - \tfrac{\eta a_t}{2} \mathbb{E}[\|\nabla L(w_t)\|^2] + \tfrac{(\eta a_t)^2 \beta \sigma^2}{2b} \right). \\
&\leq \frac{1}{T} \sum_{t=0}^{T-1} \left( \mathbb{E}[L(w_t)] + \tfrac{(\eta a_t)^2 \beta \sigma^2}{2b} \right) - \tfrac{\eta a_t}{2T} \sum_{t=0}^{T-1} \mathbb{E}[\|\nabla L(w_t)\|^2]. \\
\end{align}
Rearranging terms and noting that \(\tfrac{1}{T} \sum_{t=0}^{T-1} \big( \mathbb{E}[L(w_t)] - \mathbb{E}[L(w_{t+1})] \big) = \tfrac{1}{T} \big( L(w_0) - \mathbb{E}[L(w_T)] \big)\), we obtain
\begin{align}
\frac{\eta a_t}{2T} \sum_{t=0}^{T-1} \mathbb{E}[\|\nabla L(w_t)\|^2] &\leq \frac{1}{T} \sum_{t=0}^{T-1} \big( \mathbb{E}[L(w_t)] - \mathbb{E}[L(w_{t+1})] \big) + \tfrac{(\eta a_t)^2 \beta \sigma^2}{2b}. \\
&= \tfrac{1}{T} \big( L(w_0) - \mathbb{E}[L(w_T)] \big) + \tfrac{(\eta a_t)^2 \beta \sigma^2}{2b}.
\end{align}
Dividing both sides by \(\tfrac{\eta a_t}{2}\) gives
\begin{align}
\frac{1}{T} \sum_{t=0}^{T-1} \mathbb{E}[\|\nabla L(w_t)\|^2] \leq \frac{2}{T \eta a_t} \big( L(w_0) - \mathbb{E}[L(w_T)] \big) + \frac{\eta a_t \beta \sigma^2}{b}.
\end{align}
\end{proof}


\end{document}